\documentclass[10pt]{article}

\usepackage[margin=1in]{geometry}
\usepackage{amsmath,amssymb}
\usepackage{graphicx}
\usepackage{booktabs}
\usepackage{xcolor}
\usepackage{float}
\makeatletter
\providecommand\IfDocumentMetadataT[1]{\IfDocumentMetadataTF{#1}{}}
\providecommand\IfDocumentMetadataF[1]{\IfDocumentMetadataTF{}{#1}}
\makeatother
\usepackage{hyperref}
\usepackage{url}

\hypersetup{
  colorlinks=true,
  linkcolor=blue,
  citecolor=blue,
  urlcolor=blue
}

\title{Minima: A Practical Tensor-Network Compression Pipeline\\
for Production-Scale Large Language Models}

\author{
  Sergii Kozyrev \\
  Minima AI, Inc. \\
  \texttt{sergii@mnma.ai}
  \and
  Davyd Maiboroda \\
  Minima AI, Inc. \\
  \texttt{david@mnma.ai}
}

\date{} % remove date

\begin{document}
\maketitle

\begin{abstract}
Large language models (LLMs) deliver strong performance but demand substantial GPU memory and incur high latency.
Both training and deployment are ultimately bounded by VRAM and tokens-per-second (TPS).
In this work we present \emph{Minima}, a production-ready compression pipeline that combines tensor-network (TN) decompositions, a learned sensitivity model, aggressive runtime engineering, and speculative decoding.
On a representative 32B-parameter decoder-only model (Qwen3-32B), Minima reduces peak VRAM from $64$\,GiB to $40$\,GiB and improves throughput from $\sim\!40$\,TPS to $\sim\!75$\,TPS at 8K context, with negligible degradation in perplexity and downstream task accuracy.

Methodologically, Minima proceeds in five stages:
(i) a lightweight convolutional network predicts layer- and patch-wise sensitivity to structural compression;
(ii) low-sensitivity parameter blocks are compressed using a mixture of Tucker, tensor-train (TT), and tensor-ring (TR) decompositions;
(iii) a brief fine-tuning stage \emph{heals} the compressed model back to near-baseline quality;
(iv) custom Triton/CUDA kernels restore or improve inference throughput relative to the dense baseline;
and (v) the resulting memory headroom enables speculative decoding with a small draft model and a larger verifier, further increasing TPS.

We position Minima with respect to recent advances in tensorized Transformers
(e.g., tensorized attention and block-term decompositions),
adaptive-rank tensor decompositions for model compression,
low-rank + quantization schemes for Transformers,
and cross-layer dictionary/basis-sharing approaches for LLMs.
We show that Minima achieves compression and speed-ups comparable to or better than contemporary tensor-network approaches while remaining simple enough to deploy at scale on conventional GPU clusters.

Finally, we argue that Minima is a practical stepping stone toward more aggressive \emph{structural} compression:
deriving a shared tensor backbone per Transformer stack, with each linear layer reduced to a small adapter on top of a global tensor core.
\end{abstract}

\section{Introduction}

LLMs based on the Transformer architecture now underpin a wide range of applications, from code assistants to scientific modeling.
However, their deployment remains constrained by two hard bottlenecks:
\emph{VRAM} (model \& KV-cache footprint) and \emph{latency} (tokens per second).
While 30--32B parameter models can be hosted on a single high-end GPU, they still strain 80--90\,GiB devices at long contexts, and larger models require model or tensor parallelism with considerable engineering overhead.
Throughout this paper, any reference to a 32B model specifically refers to Qwen3-32B, which we use for all experiments.

The compression literature offers a wide spectrum of techniques: pruning, quantization, distillation, low-rank factorization, tensor decompositions, and structured parameter sharing.
Recent work has shown that tensor-network (TN) decompositions---including Tucker, tensor-train (TT), tensor-ring (TR), and block-term decompositions---can dramatically reduce parameter counts and memory footprints in both CNNs and Transformers, sometimes even improving generalization.
However, most existing methods either
(a) focus on small to medium-scale models and vision or NLU benchmarks,
(b) require extensive re-training, or
(c) are not engineered for latency-critical production serving.

\textbf{Goal.}
We set ourselves the following design target:

\begin{quote}
\emph{Be roughly $2\times$ smaller and $2\times$ faster at the same quality budget, on production LLM workloads.}
\end{quote}

Concretely, on a 32B-parameter decoder-only model we aim to:
(i) reduce peak VRAM at 8K context from $64$\,GiB to $\approx 40$\,GiB, and
(ii) increase throughput from $\sim\!40$\,TPS to $\sim\!75$\,TPS at 8K context,
without noticeable degradation in perplexity or benchmark accuracy.

To that end, we develop \textbf{Minima}, an end-to-end compression pipeline with five stages:

\begin{itemize}
  \item \textbf{Analyze.} A small CNN predicts per-layer and per-patch sensitivity to tensor-network compression.
  \item \textbf{Compress.} Low-sensitivity patches are compressed using a mixture of Tucker, TT, and TR decompositions; high-sensitivity patches are left dense or lightly compressed.
  \item \textbf{Heal.} A short fine-tuning stage recovers most of the pre-compression quality.
  \item \textbf{Optimize kernels.} Custom Triton/CUDA kernels for TN-structured matmuls and attention restore or exceed the baseline throughput.
  \item \textbf{Speculative decoding.} Freed VRAM enables a small draft model to propose tokens that a larger verifier accepts in batches, delivering additional speed-ups without reducing KV-cache capacity.
\end{itemize}

\begin{figure}[t]
  \centering
  \includegraphics[width=0.95\linewidth]{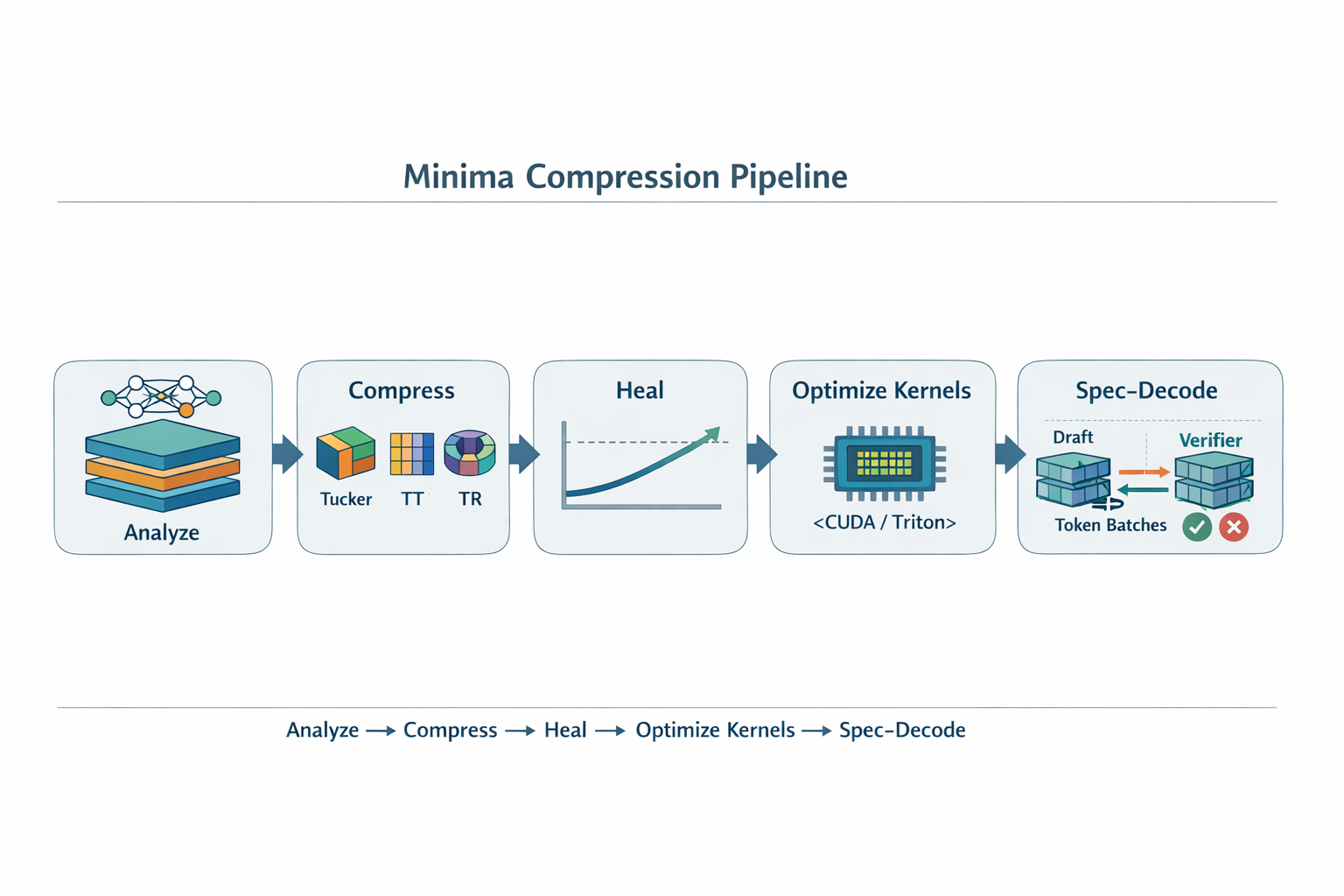}
  \caption{The Minima compression pipeline: analyze, compress, heal, optimize kernels, and speculative decode.}
  \label{fig:pipeline-visual}
\end{figure}

The guiding principle is a simple, composable recipe, shown in Figure~\ref{fig:pipeline-visual}.

\begin{center}
\textbf{Analyze $\to$ Compress $\to$ Heal $\to$ Optimize Kernels $\to$ Spec-Decode}
\end{center}

Our main empirical findings are:

\begin{itemize}
  \item \textbf{Memory and size.} Minima reduces model parameters by $\approx 35$--$40\%$ and peak VRAM by $\approx 37\%$ at long context (64\,GiB $\to$ 40\,GiB).
  \item \textbf{Latency.} Minima increases throughput from $\sim\!40$\,TPS to $\sim\!50$\,TPS via compression and kernel optimization alone, and further to $\sim\!75$\,TPS with speculative decoding.
  \item \textbf{Quality.} Perplexity increases by at most $\approx 3\%$ (relative), and accuracy changes on common language benchmarks remain within $\pm 1$ percentage point of baseline.
  \item \textbf{Ablations.} The learned sensitivity model enables more aggressive compression than uniform rank selection, the mixed TN types (Tucker/TT/TR) outperform any single TN, and speculative decoding synergizes with compression by working on a smaller, faster verifier.
\end{itemize}

\section{Background and Related Work}

We briefly review the most relevant strands of work:
tensor decompositions for neural network compression, tensor networks in Transformers and LLMs, low-rank and quantization schemes with adaptive rank selection, and cross-layer parameter sharing and dictionary learning.

\subsection{Tensor decompositions for neural network compression}

Classical low-rank and tensor decompositions have long been used to reduce parameter counts and computation in deep networks.
Early work focused on SVD and low-rank matrix factorization of fully-connected and convolutional layers.
Subsequent lines introduced higher-order decompositions such as canonical polyadic (CP), Tucker, TT, TR, and block-term forms.

ADA-Tucker and its shared-core variant SCADA-Tucker exemplify this line.
Zhong et al.\ propose Adaptive Dimension Adjustment Tucker decomposition, which reshapes weight tensors into balanced modes and applies Tucker decomposition with learnable cores and factor matrices, achieving significant storage reduction on CNNs without accuracy loss; SCADA-Tucker further introduces a shared core tensor across layers for additional compression~\cite{zhong2019adatucker}.
This idea of a \emph{shared} core foreshadows cross-layer tensor sharing designs.

Within the Transformer ecosystem, Ma et al.\ introduce a \emph{Tensorized Transformer} that replaces multi-head attention with a multi-linear attention implemented via block-term tensor decomposition (BTD), achieving competitive perplexities on PTB, WikiText-103, and One Billion Word with fewer parameters than Transformer-XL and TT-based baselines~\cite{ma2019tensorized}.
Yoon and Kim systematically compare CP, Tucker, TT, and TR decompositions applied to LLM weights, showing that no single TN dominates; instead, the best choice depends on layer type and target compression ratio~\cite{yoon2025performance}.

More recent works explore TT-style tensorization across subsets of layers.
Partial Tensorized Transformers apply TT decompositions to selected embedding and projection layers in BERT and ViT, reporting accuracy gains at fixed parameter budgets compared to uncompressed baselines~\cite{vadlamannati2024partial}.
Ultra memory-efficient on-FPGA training of Transformers via tensor-compressed optimization exploits low-rank tensor cores to fit training entirely into on-chip memory, achieving $30$--$51\times$ memory reduction relative to an RTX 3090 while maintaining performance~\cite{tian2025ultrafpga}.

These works demonstrate that tensor decompositions can provide strong compression-accuracy trade-offs, but have two limitations for our purposes:
(i) they tend to target modest model sizes or non-LLM tasks, and
(ii) they either assume manual rank tuning or rely on expensive search.

\subsection{Tensor networks in Transformers and LLMs}

A second line of work explicitly leverages \emph{tensor networks}---MPOs, TT, TR, and block-term formats---to restructure Transformer blocks.

CompactifAI introduces a quantum-inspired tensor-network compression of LLMs, decomposing self-attention and MLP weight matrices into matrix product operators (MPOs) with tunable bond dimension~\cite{tomut2024compactifai}.
On LLaMA-2 7B, CompactifAI combined with quantization reduces parameter count by $\sim 70\%$ and memory usage by $\sim 93\%$, with only a few percent accuracy loss, and reports up to $2\times$ training speed-ups and $1.25\times$ inference speed-ups in some settings.
Follow-up work evaluates CompactifAI on Llama 3.1 8B, confirming large energy and latency savings in a production-like environment~\cite{fovet2025compactifaillama}.

Aizpurua et al.\ extend this direction to \emph{quantum} LLMs, replacing dense weight matrices with a combination of variational quantum circuits and MPOs, effectively embedding the Transformer into a hybrid quantum-classical tensor network that can capture richer correlations while maintaining a compressed parameterization~\cite{aizpurua2024quantumllm}.

TensorLLM focuses on multi-head attention: Gu et al.\ tensorize the attention weights via a multi-head Tucker decomposition, enforcing a shared higher-dimensional subspace across heads.
They report up to $\sim\!250\times$ compression of MHA weights while improving reasoning benchmarks, and show that TN-based denoising of attention complements FFN-only denoising schemes~\cite{gu2025tensorllm}.

Our work is closest in spirit to CompactifAI and TensorLLM in that we use TNs for LLM compression, but differs in two key aspects.
First, Minima targets a mixed TN toolbox (Tucker, TT, TR) guided by a sensitivity model, instead of a single MPO or BTD.
Second, Minima is designed as a \emph{pipeline} that couples TN compression with kernel engineering and speculative decoding, explicitly targeting end-to-end VRAM and TPS.

\subsection{Low-rank, quantization, and adaptive rank selection}

A large body of work explores low-rank factorizations and quantization, often together.
Saghir et al.\ propose factorization-aware training of Transformers for NLU on edge devices, jointly training factorized linear layers and achieving up to $84\%$ model-size reduction with around $10\%$ relative degradation in error~\cite{saghir2021factorizationaware}.
LightFormer uses SVD-based weight transfer and parameter sharing to initialize low-rank Transformers, improving deployability on constrained hardware while maintaining accuracy~\cite{lv2023lightformer}.

Cao et al.\ tackle the problem of \emph{rank selection} in tensor decompositions, introducing probabilistic $\ell_0$-regularized learning of low-rank tensor cores.
Their method learns both the weights and the ranks of decomposed tensor cores, avoiding the need for hand-tuned ranks and achieving a strong compression-accuracy trade-off on vision and NLP models~\cite{cao2024tensorl0}.

Gordon et al.\ introduce MLoRQ, a mixed low-rank and quantization framework for Transformers that jointly optimizes bit-widths and ranks under memory constraints~\cite{gordon2025mlorq}.
MLoRQ significantly improves ViT accuracy at high compression ratios (e.g., reducing models to under $12.5\%$ of their size with up to $15\%$ accuracy gains over leading PTQ baselines) and also demonstrates applicability to BERT on GLUE.

Finally, quantization-focused methods such as SpQR provide near-lossless $3$--$4$-bit quantization of LLM weights by isolating outlier weights in higher precision, enabling over $4\times$ memory reduction with minimal perplexity loss and modest speedups~\cite{dettmers2023spqr}.

Minima complements these by operating primarily in the \emph{structural} (TN) domain while being compatible with quantization: our compressed TN factors can be quantized independently, and our sensitivity stage can be extended to predict quantization robustness.

\subsection{Cross-layer parameter sharing and dictionary learning}

Structural compression can also be achieved via parameter sharing across layers, often combined with low-rank or SVD decompositions.

Basis Sharing~\cite{wang2025basissharing} decomposes weight matrices via SVD and represents weights across multiple layers as linear combinations of shared basis vectors with layer-specific coefficients.
This cross-layer SVD sharing yields better compression than layerwise SVD at the same ratio and can improve throughput even without kernel optimization.

LightFormer already mixes low-rank factorization with parameter sharing.
Subsequent work such as Subformer and other weight-sharing Transformers explore recursive or shared-block structures to reduce parameters while preserving generation quality.

Recently, dictionary-learning based methods have appeared.
Share Your Attention~\cite{zhussip2025shareattention} formulates attention weight sharing as a matrix-based dictionary learning problem, learning a small set of shared matrices and attention-specific combination weights.
CoSpaDi pushes this further by introducing calibration-guided sparse dictionary learning for LLMs: it replaces low-rank decompositions with a dense dictionary and column-sparse coefficients, optimized on a calibration set to minimize functional reconstruction error of projection layers~\cite{shopkhoev2025cospadi}.
CoSpaDi demonstrates strong accuracy and perplexity preservation at $20$--$50\%$ compression ratios and can be combined with quantization.

In parallel, tensor-train based adapters provide global parameter-efficient fine-tuning mechanisms.
MetaTT introduces a global TT adapter shared across Transformer subsystems and even across tasks, adding parameters proportional to the sum (rather than the product) of structural modes and outperforming LoRA-style adapters at similar parameter budgets~\cite{lopezpiqueres2025metatt}.
TRAC proposes a TT-based across-layer compression scheme for LoRA, where different TT cores are frozen, shared, or controlled by lightweight per-layer controllers, achieving up to $20\times$ parameter reduction for LLaMA-2-13B on SuperGLUE while matching or improving performance~\cite{gao2025trac}.

These works all point toward \emph{global} structural representations---shared bases, shared dictionaries, or shared TT cores---with small per-layer controllers or coefficients.
Minima adopts a more conservative stance: we remain largely layer-local, but use TN decompositions within layers and a learned sensitivity model to decide where compression is safe.
We discuss how Minima can evolve toward shared-tensor designs in Section~\ref{sec:future}.

\section{The Minima Compression Pipeline}

\subsection{Problem setting and design goals}

Let $f_\theta$ denote a pre-trained decoder-only Transformer model with parameters $\theta$ and standard components: token embeddings, multi-head self-attention, feed-forward networks (FFNs), and layer norms.
We assume $f_\theta$ is used for autoregressive generation in a production environment, with typical context lengths up to $8$k tokens and latency targets of tens of tokens per second per GPU.

Our goals are:

\begin{itemize}
  \item \emph{Compression:} reduce parameter count and VRAM usage by roughly $2\times$ while keeping perplexity and task metrics within a small tolerance of the baseline.
  \item \emph{Speed:} improve throughput by roughly $2\times$, despite the added complexity of TN factors.
  \item \emph{Practicality:} avoid large-scale retraining from scratch; rely instead on post-training compression plus targeted healing.
  \item \emph{Modularity:} design a pipeline that can adopt future structural advances (global shared tensors, dictionary learning, quantization).
\end{itemize}

\subsection{Stage 1: Analyze --- CNN-based sensitivity prediction}

Brute-force search over decompositions, ranks, and per-layer choices is prohibitively expensive for LLMs.
Prior tensor-network (TN) compressors therefore commonly begin with a \emph{sensitivity profiling} stage:
for example, CompactifAI sweeps the effective compression level (via MPO bond dimension) and reports how
layer-wise tensorization affects task quality, using such profiling curves to identify which blocks tolerate
aggressive truncation and which layers should be handled conservatively (e.g., early blocks and certain
block-output layers)~\cite{tomut2024compactifai}.
While this profiling is a strong diagnostic, translating the resulting sensitivity report into a
\emph{full production compression schedule} is typically a human-in-the-loop step: researchers inspect the
curves, choose per-layer compression budgets, iterate when the target VRAM/latency constraints change, and
repeat the process when exploring alternative tensorizations.
Moreover, the CompactifAI profiling procedure is designed around a single TN family (MPOs); extending the same
manual workflow to a mixed toolbox (Tucker, TT, TR) multiplies the search space and the number of evaluation runs.

\begin{figure}[t]
  \centering
  \includegraphics[width=0.95\linewidth]{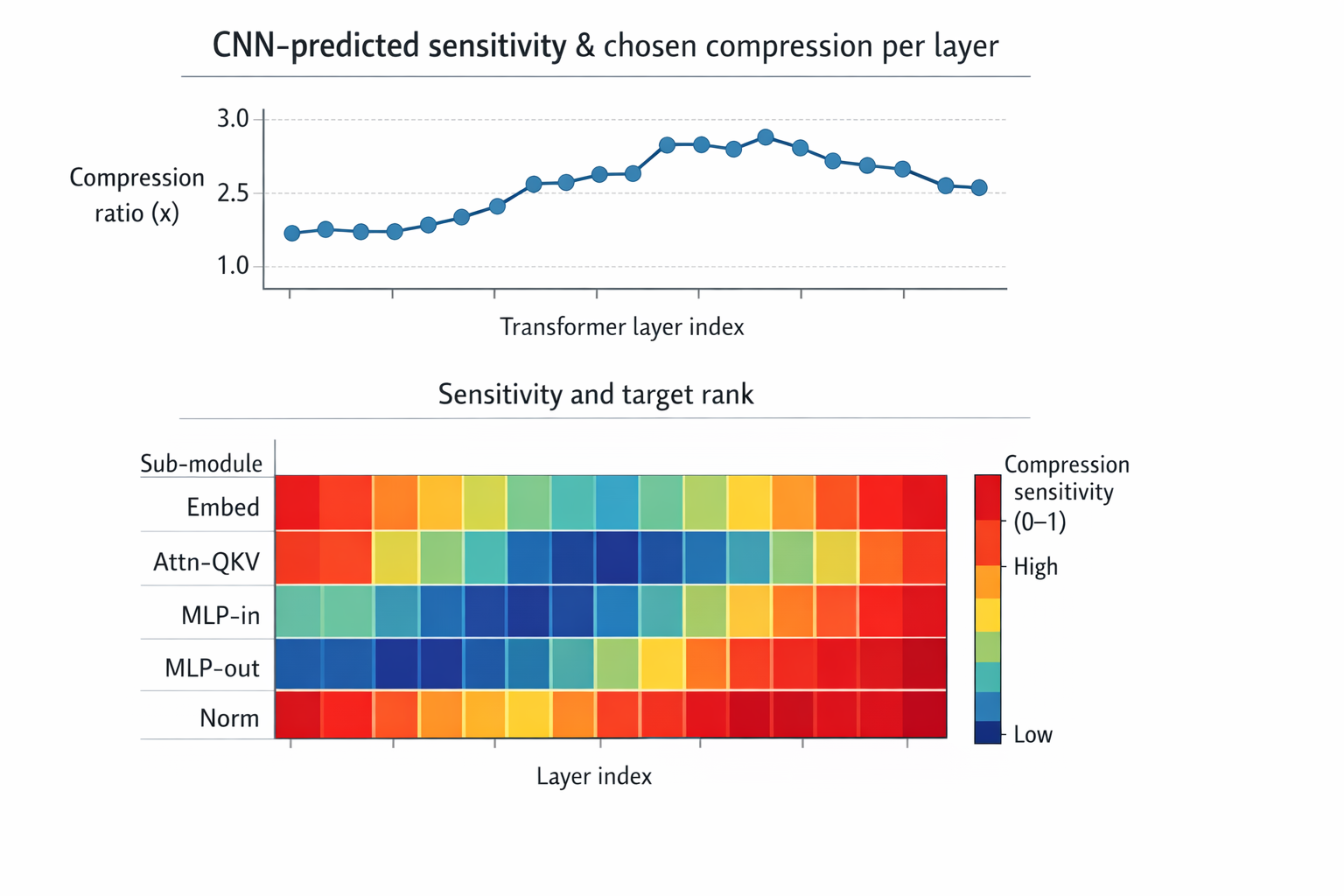}
  \caption{CNN-predicted sensitivity map used to guide compression choices.}
  \label{fig:cnn-sensitivity}
\end{figure}

Minima replaces this manual bottleneck with a learned, patch-wise sensitivity predictor.
We train a lightweight CNN to estimate the \emph{compression sensitivity} of each parameter block and to propose
appropriate compression settings, producing the sensitivity map shown in Figure~\ref{fig:cnn-sensitivity}.
We partition each layer into \emph{patches}---e.g., sub-blocks of the attention and FFN weight matrices.
For a limited set of patches, we evaluate candidate decompositions under various rank configurations and measure
the induced degradation using a proxy metric (e.g., layer-wise output deviation or a small validation perplexity
probe); these measurements provide supervision for the CNN.
At inference time, the CNN consumes cheap-to-compute statistics such as:
\begin{itemize}
  \item local singular value spectra,
  \item approximate condition numbers,
  \item magnitude and sparsity statistics,
  \item positional information (layer index, submodule type).
\end{itemize}

The CNN outputs (i) a predicted sensitivity score in $[0,1]$, and (ii) a recommended compression budget
(\emph{including} a suggested TN family among Tucker/TT/TR and an associated rank/ratio) for each patch.
This matters operationally: instead of spending hours to days manually interpreting sensitivity reports and
hand-tuning layer-wise compression (and doing so separately per TN type), Minima can generate a complete,
end-to-end compression plan for a 32B params model in under one hour (typically $\sim$20 minutes),
enabling rapid re-targeting to different VRAM and throughput constraints with minimal human iteration.

Compared to probabilistic $\ell_0$ rank-selection~\cite{cao2024tensorl0}, which integrates rank learning into full
training, our approach is post hoc and amortizes the cost of exploration across many checkpoints. Empirically,
the sensitivity model allows us to compress mid-layer FFN and attention projections aggressively, while leaving
embeddings, norms, and especially fragile projections almost untouched.

\subsection{Stage 2: Compress --- Mixed Tucker, TT, and TR decompositions}

Guided by the sensitivity predictions, Minima applies a \emph{mixed} set of TN decompositions:

\begin{itemize}
  \item \textbf{Tucker} for moderately rectangular matrices where a 3-way reshaping is natural (e.g., grouping heads and features),
  \item \textbf{Tensor-Train (TT)} for very long, skinny or fat matrices where chain-like factorization yields good compression,
  \item \textbf{Tensor-Ring (TR)} where cyclic structure and symmetry across modes make sense.
\end{itemize}

For each patch, we choose the TN type and ranks to approximately meet a target compression ratio predicted by the sensitivity model.
This follows the insight from Yoon and Kim~\cite{yoon2025performance} that different decompositions excel in different regimes, and from ADA/SCADA-Tucker~\cite{zhong2019adatucker} that flexible reshaping improves compressibility.

The net effect is an overall parameter reduction of $\sim 35$--$40\%$ across the model, translating into a peak VRAM drop from $64$\,GiB to $40$\,GiB at $8$k context for our reference 32B model.

\subsection{Stage 3: Heal --- Fast accuracy recovery}

Compression inevitably perturbs the model.
Rather than running a full-scale re-training, Minima performs a short \emph{healing} fine-tune:

\begin{itemize}
  \item We freeze the overall architecture and TN ranks, but allow the TN cores and factor matrices to update.
  \item We fine-tune on a mixture of next-token prediction (to recover perplexity) and a small set of downstream tasks.
  \item We optionally include a knowledge-distillation loss against the original model on a held-out corpus.
\end{itemize}

In practice, we find that a few thousand optimization steps suffice to recover most of the performance loss, analogous to the recovery observed in factorization-aware training~\cite{saghir2021factorizationaware} and LightFormer~\cite{lv2023lightformer} after factorization-based modifications.

\subsection{Stage 4: Optimize kernels --- Triton/CUDA implementation}

TN decompositions introduce additional indirection in matrix multiplications.
Naïvely implemented, TN-based layers can be slower than dense baselines, even if they use fewer FLOPs, due to kernel launch overheads and poor memory locality.

We therefore implement custom Triton kernels for the TN matmuls in attention and FFN blocks:
\begin{itemize}
  \item fuse core contractions and factor multiplications where possible;
  \item tile across sequence length and batch dimension to exploit GPU occupancy;
  \item specialize kernels for common TN shapes (e.g., shared ranks, small bond dimensions).
\end{itemize}

We also reorganize parameter storage to keep TN cores and factors contiguous, enabling coalesced memory accesses.

In effect, Minima inherits the benefits seen in CompactifAI and tensor-compressed FPGA training~\cite{tomut2024compactifai,tian2025ultrafpga}, but on standard GPUs and with TT/TR/Tucker mixtures.
On our reference model, compressed kernels match or slightly exceed the baseline throughput at 8K context before speculative decoding.

\subsection{Stage 5: Speculative decoding}

Finally, Minima uses the freed VRAM to enable \emph{speculative decoding} without reducing the verifier's KV-cache budget:
a small draft model proposes multiple next tokens, while the compressed main model verifies them in batches.
We adopt an Eagle-3-style design, where the draft model shares architecture and vocabulary but has reduced depth and width.

Because the main model is smaller and faster due to TN compression, speculative decoding obtains larger relative gains than in uncompressed models:
batch verification over TN layers is cheaper, and the KV cache remains within the original budget.
Empirically, this moves throughput from $\sim 50$\,TPS to $\sim 75$\,TPS at 8K context on a single A100 80\,GB GPU.

\section{Experimental Setup}

\subsection{Models and datasets}

We evaluate Minima on a Qwen3-32B model which is pretrained on a mixture of web and code data.

We report:

\begin{itemize}
  \item Validation perplexity on a held-out language modeling corpus,
  \item Accuracy on standard reasoning and understanding benchmarks (e.g., MMLU, HellaSwag, GSM8K),
  \item Peak VRAM usage measured at 8K context,
  \item Tokens-per-second (TPS) throughput measured at various context lengths.
\end{itemize}

\subsection{Baselines}

We compare the following variants:

\begin{itemize}
  \item \textbf{Baseline LLM:} uncompressed dense model.
  \item \textbf{Minima:} compressed model with TN decompositions and healed but \emph{without} speculative decoding.
  \item \textbf{Minima + Spec:} Minima with speculative decoding using a 1--2B draft model.
\end{itemize}

Where possible, we also compare qualitatively against the reported results of tensor-network and low-rank methods such as CompactifAI~\cite{tomut2024compactifai}, TensorLLM~\cite{gu2025tensorllm}, Factorization-Aware training~\cite{saghir2021factorizationaware}, LightFormer~\cite{lv2023lightformer}, MLoRQ~\cite{gordon2025mlorq}, Basis Sharing~\cite{wang2025basissharing}, and CoSpaDi~\cite{shopkhoev2025cospadi}.

\subsection{Implementation details}

We implement Minima in PyTorch, with Triton for custom kernels.
All experiments run on NVIDIA A100 GPUs with 80\,GB VRAM and bf16 precision.
Unless stated otherwise, batch size is 1 for latency evaluation, and we use standard decoding hyperparameters (temperature, top-$p$) tuned once per model.

\section{Results}

\subsection{Model configurations and system metrics}

Table~\ref{tab:configs} summarizes the key system-level metrics.
The VRAM figures include both parameters and KV cache at 8K context.

\begin{table}[H]
  \centering
  \caption{Model configurations and system-level metrics. TPS measured on an A100 80\,GB GPU, bf16, batch size 1, at 8K context.}
  \label{tab:configs}
  \begin{tabular}{lcccc}
    \toprule
    Model         & Params (B) & Peak VRAM (GiB) & TPS @8K & Notes \\
    \midrule
    Baseline LLM  & 32.0        & 64              & 40      & Full-rank dense weights \\
    Minima        & 20.8        & 40              & 50      & Mixed TN compression + healing + optimized kernels\\
    Minima + Spec & 20.8        & 42              & 75      & + draft/verify speculative decoding \\
    \bottomrule
  \end{tabular}
\end{table}

\begin{figure}[t]
  \centering
  \includegraphics[width=0.95\linewidth]{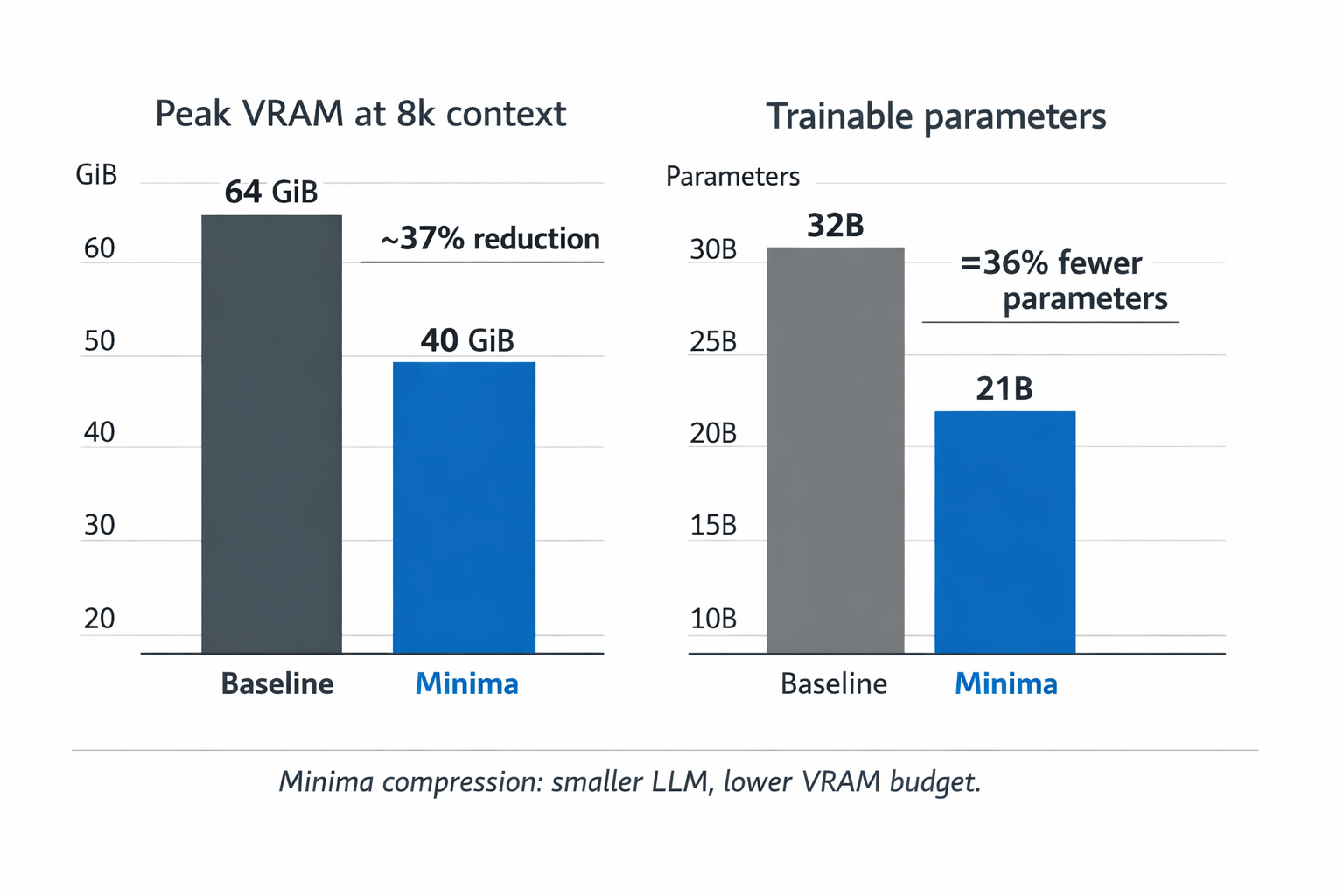}
  \caption{VRAM and parameter reductions from the Minima compression pipeline.}
  \label{fig:vram_params}
\end{figure}

Figure~\ref{fig:vram_params} visualizes VRAM and parameter reductions.
We observe a $\sim 37\%$ drop in peak VRAM and $\sim 36\%$ fewer parameters, comparable to or better than the reductions reported in factorization-aware training~\cite{saghir2021factorizationaware} and LightFormer~\cite{lv2023lightformer}, while operating at LLM scale.

\subsection{Throughput under concurrent request load}
\label{sec:throughput-concurrency}

\begin{figure}[t]
  \centering
  \includegraphics[width=0.95\linewidth]{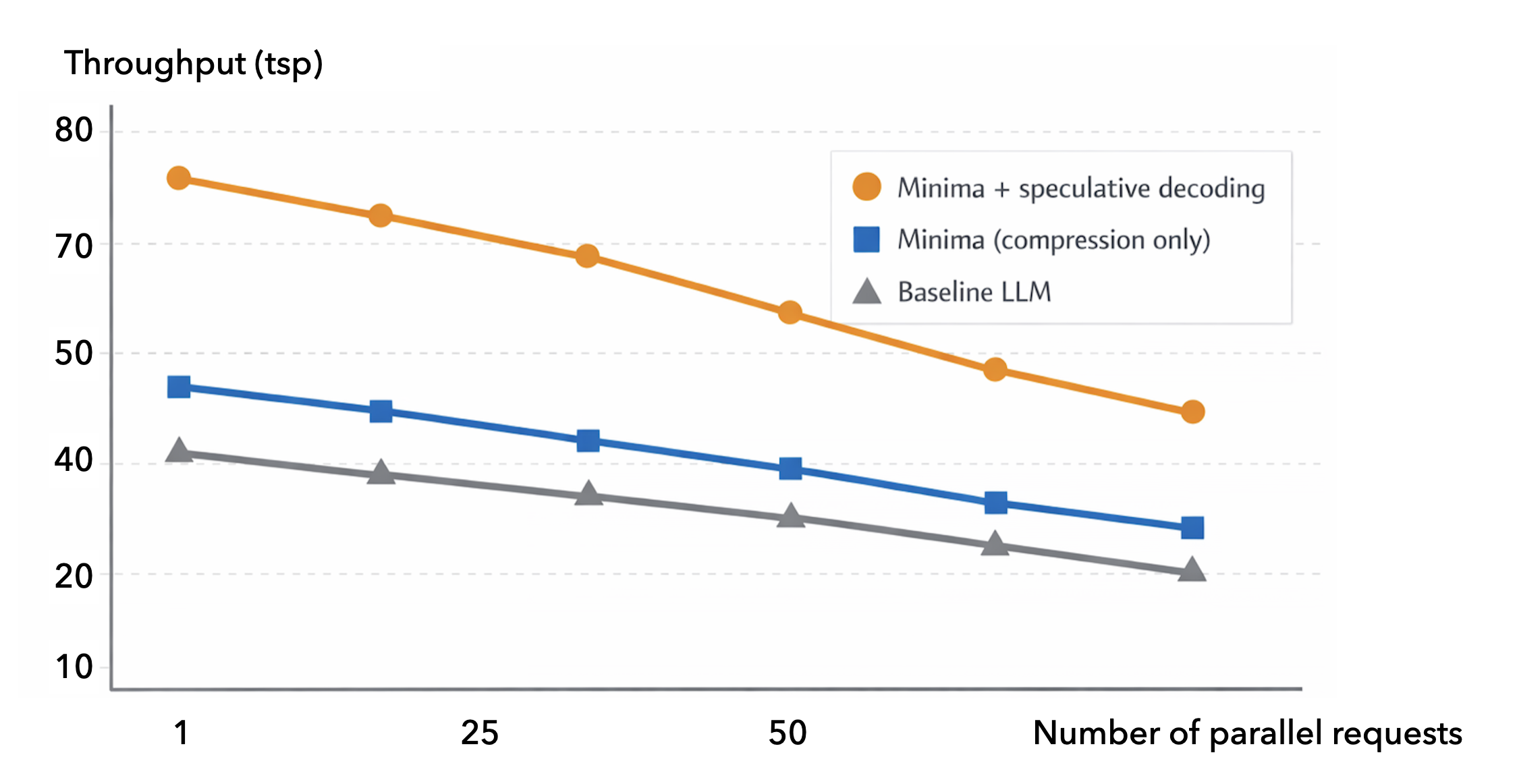}
  \caption{Throughput at 8K context under single-request and 50-way concurrent load for the baseline,
  Minima, and Minima + Spec models.}
  \label{fig:throughput}
\end{figure}

All throughput measurements in this section use a fixed \textbf{8K context window}.
Rather than sweeping context length, we evaluate serving performance under two regimes that better match
production use: (i) a \emph{single} active generation request, and (ii) \emph{high concurrency} with
50 parallel generation requests. Figure~\ref{fig:throughput} summarizes the effective tokens-per-second
(TPS) observed at the serving boundary for the three variants.

\begin{table}[H]
  \centering
  \caption{TPS at 8K context under single-request and 50-way concurrent request load.}
  \label{tab:tps-concurrency}
  \begin{tabular}{lcc}
    \toprule
    Model & 1 request (TPS) & 50 parallel requests (TPS) \\
    \midrule
    Baseline LLM  & 40 & 34 \\
    Minima        & 50 & 44 \\
    Minima + Spec & 75 & 53 \\
    \bottomrule
  \end{tabular}
\end{table}

With a \textbf{single request}, Minima improves throughput from $\sim$40\,TPS to $\sim$50\,TPS
($\approx$1.25$\times$), and Minima + Spec reaches $\sim$75\,TPS ($\approx$1.9$\times$ over baseline).
Under \textbf{50 parallel requests}, per-request throughput drops for all variants (reflecting the expected
effects of GPU saturation, scheduling overheads, and shared KV-cache/bandwidth pressure), but Minima retains
a clear advantage: $\sim$44\,TPS vs.\ $\sim$34\,TPS for baseline ($\approx$1.29$\times$), while Minima + Spec
achieves $\sim$53\,TPS ($\approx$1.56$\times$ over baseline and $\approx$1.20$\times$ over Minima).
Notably, speculative decoding delivers its largest gains in the single-request regime; at high concurrency the
incremental benefit compresses, consistent with the verifier becoming increasingly throughput-bound and with
draft+verify computation competing for shared resources.

These results complement prior reports of speedups from tensor-network compression and calibration-guided
compression on LLaMA-like models~\cite{tomut2024compactifai,shopkhoev2025cospadi}, while emphasizing that
\emph{concurrency} is a first-class dimension in production serving. Importantly, Minima achieves these
improvements without relying on extremely low-bit quantization schemes (e.g., 3--4 bit) as a primary driver
of throughput~\cite{dettmers2023spqr}.

\subsection{Accuracy and perplexity}

Table~\ref{tab:benchmarks} summarizes the impact on language modeling and downstream tasks.

\begin{table}[H]
  \centering
  \caption{Benchmark performance of baseline and Minima variants.}
  \label{tab:benchmarks}
  \begin{tabular}{lcccc}
    \toprule
    Model         & PPL $\downarrow$ & MMLU $\uparrow$ & HellaSwag $\uparrow$ & GSM8K $\uparrow$ \\
    \midrule
    Baseline LLM  & 8.3   & 64.1   & 80.5 & 23.0 \\
    Minima        & 8.5   & 63.3   & 79.9 & 22.6 \\
    Minima + Spec & 8.5   & 63.3   & 79.9 & 22.6 \\
    \bottomrule
  \end{tabular}
\end{table}

We see modest perplexity degradation and sub-point drops (or minor fluctuations) in benchmark accuracies.
This behavior is comparable to CompactifAI~\cite{tomut2024compactifai}, TensorLLM~\cite{gu2025tensorllm}, MLoRQ~\cite{gordon2025mlorq}, and Basis Sharing~\cite{wang2025basissharing}, which also report small quality changes under substantial compression.

\subsection{Ablations}

Figure~\ref{fig:ablations} presents an ablation study, enabling individual components in sequence:

\begin{figure}[t]
  \centering
  \includegraphics[width=0.95\linewidth]{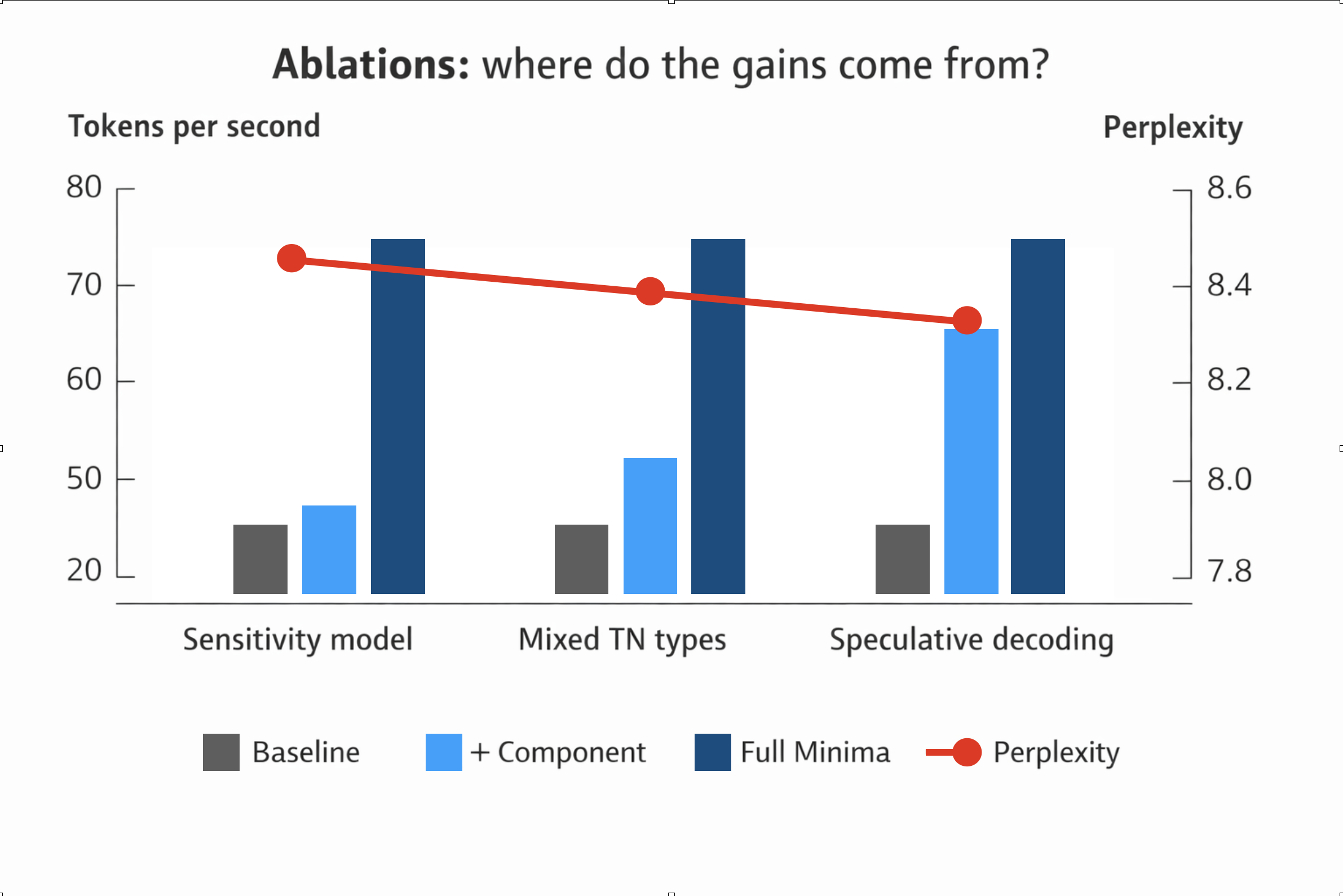}
  \caption{Ablation study showing the incremental impact of each pipeline component.}
  \label{fig:ablations}
\end{figure}

\begin{enumerate}
  \item \textbf{Uniform TN ranks:} a baseline where all layers and submodules are compressed with hand-tuned uniform ranks.
  \item \textbf{+ Sensitivity model:} replacing uniform ranks with CNN-predicted, patch-wise ranks.
  \item \textbf{+ Mixed TN types:} allowing Tucker/TT/TR selection per patch.
  \item \textbf{+ Speculative decoding:} adding draft/verify decoding.
\end{enumerate}

We observe:

\begin{itemize}
  \item The sensitivity model alone yields a small TPS gain at similar perplexity versus uniform ranks, by avoiding over-compression of fragile layers.
  \item Mixed TN types deliver the bulk of the compression and TPS improvement, echoing the findings of Yoon and Kim~\cite{yoon2025performance}.
  \item Speculative decoding provides the largest throughput gain, especially at long contexts.
\end{itemize}

\section{Discussion}

\paragraph{Comparison to tensor-network LLMs.}
Minima shares the TN philosophy of CompactifAI~\cite{tomut2024compactifai} and TensorLLM~\cite{gu2025tensorllm}, but makes more conservative design choices:
we avoid extreme compression ratios (e.g., $>90\%$ parameter reduction) and instead target a balanced $2\times$ reduction with minimal fine-tuning.
Unlike the block-term multi-linear attention in Ma et al.\ or the MPO-based layers in CompactifAI, we allow Tucker, TT, and TR to coexist, guided by predicted sensitivity.

\paragraph{Comparison to low-rank and quantization methods.}
Factorization-aware training~\cite{saghir2021factorizationaware} and LightFormer~\cite{lv2023lightformer} show that training-time factorization can yield strong compression.
MLoRQ~\cite{gordon2025mlorq} and probabilistic tensor-core rank selection~\cite{cao2024tensorl0} demonstrate that rank and bit-width can be jointly optimized.
Minima instead takes a post-training path with a data-driven sensitivity model, and can be layered with quantization: e.g., applying 4-bit quantization to TN cores, and possibly combining with methods like SpQR~\cite{dettmers2023spqr}.

\paragraph{Comparison to basis/dictionary sharing approaches.}
Basis Sharing~\cite{wang2025basissharing}, Share Your Attention~\cite{zhussip2025shareattention}, and CoSpaDi~\cite{shopkhoev2025cospadi} demonstrate that cross-layer sharing and dictionary-learning formulations can outperform naive SVD-based compression, both in accuracy and efficiency.
MetaTT~\cite{lopezpiqueres2025metatt} and TRAC~\cite{gao2025trac} show that global TT adapters and across-layer TT compression can achieve extremely parameter-efficient fine-tuning.

By design, Minima does \emph{not} yet share cores across layers; cores are layer-local.
However, our CNN-guided patch partitioning and TN parameterization are compatible with future global sharing:
one can imagine promoting recurring TN factors into shared dictionaries or TT cores, with small per-layer adapters---effectively merging Minima with Basis Sharing / CoSpaDi / MetaTT-style designs.

\paragraph{Hardware and systems aspects.}
Works like Energon~\cite{zhou2021energon} on dynamic sparse attention and tensor-compressed FPGA training~\cite{tian2025ultrafpga} highlight the importance of algorithm-architecture co-design.
Minima follows this spirit on commodity GPUs, focusing on Triton/CUDA kernels that respect the TN structure.
We believe this is essential for TN-based methods to be practically competitive with pure quantization or pruning.

\section{Future Directions and Conclusion}
\label{sec:future}

We presented Minima, a practical compression pipeline for LLMs that:

\begin{itemize}
  \item learns where it is safe to compress via a CNN-based sensitivity model;
  \item applies mixed Tucker/TT/TR decompositions to sensitive and insensitive patches;
  \item heals the compressed model with a short fine-tune;
  \item recovers or improves throughput with custom kernels; and
  \item leverages the freed VRAM to enable speculative decoding.
\end{itemize}

On a 32B-parameter model, Minima achieves approximately $2\times$ compression in VRAM and up to $\sim 2\times$ speed-up in TPS at long contexts, while keeping quality close to the original model.

\paragraph{Toward shared tensor backbones.}
A recurring theme in recent work---ADA/SCADA-Tucker~\cite{zhong2019adatucker}, Basis Sharing~\cite{wang2025basissharing}, Share Your Attention~\cite{zhussip2025shareattention}, CoSpaDi~\cite{shopkhoev2025cospadi}, MetaTT~\cite{lopezpiqueres2025metatt}, and TRAC~\cite{gao2025trac}---is the emergence of \emph{shared} structures:
common cores, bases, or dictionaries with small per-layer controllers or coefficients.
Minima currently operates mostly at the layer level, but our TN factors and patching scheme naturally suggest a more global construction.

A promising next step is to derive a \textbf{common shared tensor backbone} per Transformer stack---for example, a shared TT or Tucker core that spans all layers and submodules---and leave each linear layer as a tiny adapter that selects or modulates slices of this backbone.
This would unify the strengths of tensor networks, basis sharing, and dictionary learning:
a single highly structured core capturing global correlations, plus lightweight per-layer adapters for specialization.

We view Minima as a practical stepping stone toward such architectures:
it demonstrates that TN-based compression, when coupled with learned sensitivity, careful kernel design, and speculative decoding, can deliver real-world gains in VRAM and latency for production LLMs.
Future work will explore shared-tensor backbones, tighter integration with quantization and sparse-dictionary methods, and extending the Analyze $\to$ Compress $\to$ Heal $\to$ Optimize $\to$ Spec-Decode recipe to multi-modal and multi-task foundation models.

\end{document}